# Probability Estimation in face of Irrelevant Information

**Adam J. Grove**
Department of Computer Science
Stanford University
Stanford, CA 94305
grove@cs.stanford.edu

**Daphne Koller**
Department of Computer Science
Stanford University
Stanford, CA 94305
daphne@theory.stanford.edu

## Abstract

In this paper, we consider one aspect of the problem of applying *decision theory* to the design of agents that learn how to make decisions under uncertainty. This aspect concerns how an agent can estimate probabilities for the possible states of the world, given that it only makes limited observations before committing to a decision. We show that the naive application of statistical tools can be improved upon if the agent can determine which of his observations are truly relevant to the estimation problem at hand. We give a framework in which such determinations can be made, and define an estimation procedure to use them. Our framework also suggests several extensions, which show how additional knowledge can be used to improve the estimation procedure still further.

## 1 INTRODUCTION

The problem we consider in this paper is how to estimate probabilities for states of the world, so that agents can use the techniques of *decision theory* to make decisions under uncertainty. We illustrate this problem with an example. Suppose we wish to design an agent $M$ whose function is to deliver packages around town as swiftly as possible. We could program $M$ with a set of different methods for doing this; for instance, it can drive between destinations using either the freeway system or city roads, it can walk, it can give the package to the postal service to deliver instead, and so on. Let us ignore the enormous task of implementing each method, which is, in essence, a planning problem and beyond the scope of this work. Here we ask how the agent is to decide between them.

For this example, we might take the following simplistic view of the world. First, we suppose that the time taken to drive depends, in a known way, only on whether traffic is congested. The time taken to walk depends (again, in a known way) on the weather conditions. Posting the package takes constant time. Now we ask $M$ to deliver a particular package. $M$ must commit to a method before finding out about the traffic or whether it will rain. If $M$ can associate a probability with each relevant possibility, then it can calculate expected time for each method and decide accordingly.

We will assume that $M$ has been in situations like this before (presumably this is the case once it has been operational for a while), and so it can estimate probabilities from stored observations. However, $M$ might actually know a lot about the current situation: it could know the time, the day of week, the season, today's weather forecast, the package's weight, the recipient, whether the car has been serviced recently, the price of a postage stamp, and so on. What we really want is the probability of each of the events of interest (such as, it is sunny but traffic is light) conditioned on what is known. This presents a problem because, once we take all of this knowledge into account, most of $M$'s previous data no longer pertains directly. The main issue studied in this paper is how we can decide what information can be safely ignored as being irrelevant, so improving the quality of the estimated probabilities. The agent $M$ might have very little data directly applicable to estimating the chance of "fine weather and light traffic" given all he knows, because it is likely $M$ hasn't been in an identical situation often before. But we know that a lot of $M$'s knowledge—for instance, the nature of the package and the condition of the car—has nothing at all to do with the weather or road conditions. If we estimate probabilities conditional on just relevant data, such as the forecast and the time of day, there will be many more observations that can be used.

The model which stands at the heart of classical decision theory, and on which our work is based, is the decision matrix (see [Savage, 1954] for definitions, and Section 5 for some discussion of related work in AI). As explained in Section 2, applying this technique requires the estimation of the probabilities of the various states of the world that the agent considers possible. Using these probabilities, the agent can estimate the expected utility for each alternative action, and choose the one that maximizes that quantity. Unlike many previous works (e.g. [Simon and Kadane, 1975]), we do not assume that these probabilities are made available by an external source. More realistically, we assume that the agent uses its own experience as the major source of information. The agent will thus learn from experience, by gradually refining its estimates.

Our method for estimating probabilities, which is based on a procedure that attempts to discover irrelevant attributes, is described in Section 3. Some extensions are outlined in Section 4. Our technique combines concepts reminiscent of probabilistic reference hierarchies (see, for example, [Bacchus, 1988]) with statistical tools. It thus enables using statistical data, as well as less precise notions about relevance that the designer might have.

## 2    THE UNDERLYING MODEL

The decision-making module takes a decision problem, creates a decision-matrix, and uses that matrix to decide on a course of action. Each row of the decision matrix is a possible action under consideration by the agent. For example, agent $M$'s actions might consist of: drive on a freeway, drive on the city roads, walk, and send by mail. Of course, these high-level actions usually represent complex plans consisting of many atomic steps; the model we are using ignores the planning problem of how to determine these steps. The columns of the matrix are possible states of the world, where each state has an associated probability (the probability that it holds in this particular decision situation). These two components of the matrix will be described later in this section. The elements of the matrix are the agent's outcomes for each action/state pair. It is a well-known result of decision theory (see [Savage, 1954]) that in many cases, the agent's preference ordering on outcomes can be expressed by numerical *utilities*. In the particular context of intelligent agents, the utility will often be expressed in terms of certain parameters, such as time, fuel consumption, or money. For example, we might use time as our measure of utility for $M$. In our paper, we assume that the matrix entries are utilities, and are given in advance. Once the entire decision matrix is available, the agent simply chooses the action with maximum expected utility.

### 2.1    STATES AND EVENTS

We adopt a framework in which the agent observes and reasons about the world using a fixed set of *attributes*, $\mathcal{A} = \{A_1, \ldots, A_n\}$, which take on values in a finite space. For example, the attribute "day of the week" has a natural set of values; an attribute such as "weather" would be partitioned into values (e.g. raining, cloudy, fine), where the granularity of the partition will depend on the agent's needs. This vocabulary must be chosen carefully, as it greatly affects the performance of the decision-making module The attributes should be chosen to be, in some sense, independent of each other. See Section 4 for further discussion.

The most specific assertion we can make about the world is to announce the value of each $A \in \mathcal{A}$. This exactly determines the world as far as the agent's vocabulary allows it to differentiate. We therefore define a *state* to be an assignment of a value to every attribute in $\mathcal{A}$. In the decision-theoretic paradigm, the agent's uncertainty is modeled via the existence of several states that the agent believes the world could be in.[1] In general, some of the attributes in $\mathcal{A}$ will have no connection to the decision at hand (i.e. will not affect the outcomes of the contemplated actions). For example, although $M$ might have an attribute describing the last time the car was serviced, the value of this attribute would not be relevant to the time it takes $M$ to deliver a certain package. We define an *event* $E$ to be a list of values for a subset $\mathcal{A}'$ of the attributes in $\mathcal{A}$. We say that an event $E$ *obtains* when the true state agrees with $E$ on all attributes in $\mathcal{A}'$.

We assume that the columns of the decision matrix are all events which contain values for some specific subset $\mathcal{A}'$ of $\mathcal{A}$. Ideally, these attributes should be those that have some connection with the actions or the decision under consideration. The smaller $\mathcal{A}'$ is, the easier it is to make the decision.[2] In our example, $\mathcal{A}'$ might consist of the attributes "weather" and "traffic density", and the column events would be all the possible assignments of values to these two attributes.

The initial information $I$ denotes the set of attribute-value pairs that the agent observes in a particular decision making situation, i.e., before an action is chosen and executed. It should be clear that $I$ is also an event.

### 2.2    PROBABILITY DISTRIBUTION

The decision-theoretic model we are using here assumes the existence of some objective probability distribution on the possible states. That is, let $\mathcal{W}$ be the set of all possible states of the world. We are assuming that $\mathcal{W}$ has the additional structure of a (presumably unknown) probability space $(\mathcal{W}, \pi)$. It is unlikely to be true, in any meaningful sense, that the actual state of the world is a random draw from some probability distribution, but the probabilistic model is often a good approximation to the intricately causal way the world actually works. For further discussion, see, for example, [Cox, 1961; Jaynes, 1968; Savage, 1954].

---

[1]This is similar to the familiar concept of "possible worlds".

[2]For choosing $\mathcal{A}'$, all we need is some, possibly incomplete, knowledge about which attributes are "relevant" to an action. It is not necessary to know how or why a certain attribute affects the consequence of the action, or even be certain that it really does.

In our model, a decision-making situation evolves as follows. At the time the agent learns that he is to make a decision, the actual state of the world is regarded as being randomly chosen according to $(\mathcal{W}, \pi)$. The agent has the capability of observing the values of some of the attributes, and so gains some initial information $I$ about the chosen state. It then chooses an action (i.e., makes a decision), and afterwards, perhaps because of the action's execution, it learns more about what the world was like at decision time. That is, it learns the values of more attributes.

## 2.3  DATA COLLECTION

We assume that the agent has a database $\mathcal{D}$ of observations relating to past experiences (the integration of other types of data into our model will be discussed briefly in Section 4). To be more specific, we assume that every data point $D \in \mathcal{D}$ actually arose from some earlier decision-making episode. Thus $D$ contains, among possibly other things, the values for the attributes that were observed before and during the decision process, i.e., the agent's information about the state of the world holding at the time.

In order to simplify the model, we might make a *complete observability* assumption: the agent always observes the value of every attribute in $\mathcal{A}$. Consider the following example, which illustrates what can go wrong without some such requirement. Suppose that $M$ is sometimes told about baseball games taking place in the city, and that when there is a baseball game $M$ usually decides to walk (perhaps because baseball games generally take place when the weather is fine). Suppose, in violation of complete observability, that if the agent chooses to walk it does not find out about the traffic conditions. Then if the traffic density is in fact heavier on days in which a game takes place, $M$'s estimate for the probability of heavy traffic will be lower than the true value, because most of its observations will be taken on days when there is no game. In general, complete observability avoids this and similar errors because it implies that every observation in $\mathcal{D}$ truly is a random sample from $(\mathcal{W}, \pi)$, free of unwanted bias.

In practice, complete observability can be weakened to *independent observability*, which says that the set of attributes the agent gets to learn about is determined independently of both the actual state of the world, and of any decision the agent takes. In the above example this was violated, because whether or not the traffic was observed depended on whether the agent decided to walk, and it was this that induced bias. Of course, the independent observability requirement by itself offers no guarantee that we ever see enough data to estimate all the required probabilities. So it is also necessary to assume that, whenever $\mathcal{A}'$ defines the set of events for some decision the agent might be asked to make, then the chance of observing this set should be nonzero. Even this requirement can be weakened. For instance, if two attributes are rarely observed together, appropriate assumptions about conditional independence can be used so as to still permit accumulation of sufficient historical data.

In subsequent sections, we shall simplify the presentation by stating our results and techniques in terms of the complete observability assumption only. However, the extensions to the weaker, but more realistic, conditions are straightforward.

In practice, the most restrictive consequences of our model are as follows. First, the requirement that the information observed is an event amounts to assuming that the agent either identifies, without uncertainty, the value of an attribute or else learns nothing at all about it. But we note that if the set $\mathcal{A}$ is chosen well, this assumption should cause relatively little difficulty. The other problem with our model relates to the amount of data stored: there is little obvious scope for data summarization or compression. Currently, our estimation procedure requires every observation to be remembered (or, only slightly better, remember counts for observations that occur frequently). Significant improvements are likely to depend on domain-specific structure.

## 3  THE ESTIMATION PROBLEM

Let us review the probability estimation problem. We have some initial information $I$. We have determined a list of events which are the columns of the decision matrix, and wish to estimate $p = Pr(E|I)$ for each event $E$. Of course, how best to form these estimates is a problem in statistics.

One simple and theoretically sound estimate of $p$ is simply the proportion of data points agreeing with $E$, among all points that agree with $I$. This estimate is "good" in several ways: for instance, it is unbiased (i.e., the expected value is exactly $p$) and its variance decreases to zero as the number of relevant data points grows. The problem is that the number of relevant data points may not grow very quickly because this estimate uses only those observations which agree *exactly* with $I$. But perhaps situations matching $I$ have not been encountered very often. For instance, if time and date are part of $I$, there will be no relevant historical data at all. We conclude that this simple estimation procedure is often impractical.

To salvage the approach to decision making we are looking at, we need to be much more clever about how we estimate probabilities. The main result of this section is a technique for probability estimation which can yield substantially better results than the above. It does this by providing a framework which can capture and make use of additional information we have about the structure of the world (see Section 4). The underlying idea is the observation that the less specific the information in $I$, the more useful data points we will have. Therefore, the estimate would improve if we could (justifiably!) ignore some of the attributes mentioned in $I$.

It seems to be often true that, in any given context, only a few attributes will be *relevant* to whether some event $E$ occurs. Consider the example in the introduction, where $I$ records, among other things, which day of the week it is. If the attributes in $E$ all refer to natural phenomena, such as the weather conditions, we would expect the day of the week to be irrelevant to—and, in a sense which is easy to make precise, *independent* of—whether $E$ occurs. In this case, the best estimate of $p$ would pool data for all days, even though this ignores some of the knowledge contained in $I$. In general, it is not reasonable to require all such information about relevance to be supplied ahead of time. In the example, we would want the estimation procedure to find out for itself whether the day of the week is unimportant.

We begin by looking at the base case of our technique. Suppose $I$ includes the value of some attribute $A$. Let $I_1, I_2, \ldots, I_k$ be all events which are just like $I$, except possibly with respect to the value of $A$; we may assume that $I_1 = I$. Intuitively, we can pool data only if our knowledge about $A$ is irrelevant to $p$. More formally, we ask whether the conditional probabilities $p_i$ (i.e., $Pr(E|I_i)$) are the same for all $i$. Our procedure is to test whether this independence is plausible, and then use either the pooled or non-pooled estimate as appropriate. Both the test and the estimate itself can make use of all the observed data in $\mathcal{D}$.

In the following, let $N_i$ be the number of observations in $\mathcal{D}$ agreeing with $I_i$. Let $\hat{p}_i$ be the proportion of these observations that do in fact agree with the values specified in $E$. We estimate $p$ either as $\hat{p}_1$ or else as the pooled estimate $\hat{p} = (\sum_{i=1}^{k} \hat{p}_i N_i)/N$, where $N = \sum_{i=1}^{k} N_i$. Note that $\hat{p}_1$ is simply the direct estimate which was mentioned earlier. We decide which of these two possibilities to use on the basis of a hypothesis test (the hypothesis being that $p_i = p_j$, for all $i, j$.)

One relatively simple hypothesis test we can use for this is the $\chi^2$ test, which is discussed in most statistics texts (such as, [Larsen and Marx, 1981; Sachs, 1982]). The test is based on the value $X^2 = \sum_{i=1}^{k} (\hat{p}_i - \hat{p})^2 N_i/(\hat{p}(1-\hat{p}))$. If the hypothesis (equal $p_i$) is true, and the $N_i$ are not too small,[3] then the dis-tribution of $X^2$ is very well approximated by the $\chi^2$ distribution with $k-1$ degrees of freedom. In order to perform the test, we must choose some small $\alpha > 0$, which becomes the chance of not pooling data when it really would have been permissible. We accept the hypothesis and use the pooled estimate just if $X^2 < c_\alpha$, where $c_\alpha$ is such that the chance of a random sample from $\chi^2$ exceeding $c_\alpha$ is $\alpha$. The value $c_\alpha$ can be found from tables. It is generally desirable to have $\alpha$ very small, but note that as $\alpha$ decreases the chance of incorrectly deciding to pool data when this is not justified grows. Later we state two asymptotic properties of our estimation procedure, whose proof assumes that $\alpha$ is $1/N^d$, for some $d > 1$. That is, as the sample size increases we should tolerate less chance of deciding incorrectly not to pool data. It turns out that, so long as $\alpha$ grows smaller no faster than this, the chance of pooling inappropriately also diminishes rapidly.

To recap, the general idea of our procedure is to test for independence and then use the estimate suggested by the result of the test. The hypothesis test can be done in many ways, and we have suggested one possibility, the $\chi^2$ test. We chose this test because its simplicity facilitates the analysis.

This analysis is important because the procedure uses the same data for both the independence test and for the actual estimate. In this way, we can hope to make the best use of scarce data. But reusing sampled data destroys the independence between the outcome of the hypothesis test and the estimate used, and so we must check that the process as a whole gives us a useful result. For example, it is easy to see that both the pooled estimate $\hat{p}$ and the simple estimate $\hat{p}_1$ are unbiased if the hypothesis of equal $p_i$ is in fact true. But it does not follow just from this that the composite estimate is unbiased.[4] Another related issue concerns the estimate's variance: intuitively, we only pool data if all the $\hat{p}_i$ are approximately the same, and so it might seem that the pooled estimate is only used when it provides little additional information over $\hat{p}_1$ anyway.

It turns out that neither of these problems arise: the estimate we give is asymptotically unbiased, and has asymptotic variance that can be much smaller than $\hat{p}_1$. In other words, at least for large $N$, our estimate is indeed very likely to be close to the true value. Furthermore, if the hypothesis of equal $p_i$ is true, then our estimate has smaller variance than the simple unpooled estimate $\hat{p}_1$ and so is likely to be much closer to $p$. The formal statement of these results is contained

---

[3] A frequently stated rule of thumb is that $N_i p_i$ should be larger than 5, for all $i$.

[4] To illustrate the possible problems, suppose that the test is such that the hypothesis of equal $p_i$ is slightly more likely to be accepted when the observations satisfy $\hat{p} < \hat{p}_1$ than it is otherwise. Then this would bias our estimate. Because we reuse data, the test does get to see the actual values of the estimates $\hat{p}$ and $\hat{p}_1$, and so such behavior cannot be ruled out without deeper analysis.

in the following two theorems (whose proofs are too long for inclusion here).[5]

**Theorem 3.1:** *If in fact $p_i = p$ for all i, then the estimate we give has mean $\mu$ and variance $\sigma^2$ such that $\mu \longrightarrow p$ and $\sigma^2 \longrightarrow p(1-p)/N$ as $N \longrightarrow \infty$.[6] We note that this asymptotic variance is the best that can be achieved by any unbiased estimate, even if we know for certain that the hypothesis of equal probabilities is true.*

**Theorem 3.2:** *If in fact $p_i \neq p_j$ for some $i, j$ then the estimate we give has mean $\mu$, and variance $\sigma^2$ such that $\mu \longrightarrow p_1$ and $\sigma^2 \longrightarrow p_1(1-p_1)/N_1$ as $N \longrightarrow \infty$. We note that this asymptotic variance is the smallest possible amongst unbiased estimates of $p_i$ (given that observations relating to $I_i$, for $i \neq 1$, are regarded as being not informative about $p_1$).*

Although these theorems give asymptotic results only, it seems very likely that the procedure will work well for far smaller sample sizes than were required in our proof. Proving a precise claim about this would be difficult. Instead, we have programmed the technique to run on simulated data, and the results there did confirm this expectation. When the data was generated for each class using the same underlying probability, the decision was made to pool data most of the time. In one typical experiment, 200 data points were successively generated for each of five events, and on average the procedure declined to pool data less than 5% of the time. If the probabilities differ between classes, even by relatively small amounts—and note that the closer the probabilities are to being equal, the less damage is done by incorrect pooling—our procedure rapidly discovered this. In one experiment, where the difference between all probabilities was less than or equal to 0.1, the procedure apparently stabilized on a decision not to pool after each class had accumulated about 150 data points. A similar experiment where the probabilities differed by up to 0.3 stabilized after about 15 data points per event on average.

The procedure so far will, in effect, decide whether to ignore one particular attribute of $I$. In general, several attributes of $I$ may turn out to be irrelevant. Our technique extends easily to such cases. Suppose that we have decided to ignore some attribute $A$ of $I$ (using a test like that just suggested). That is, we have accepted the hypothesis that $Pr(E|I) = Pr(E|I')$ for all

$I'$ that are like $I$ except for the value of $A$. But from this it also follows that $Pr(E|I) = Pr(E|(I-A))$, where by $I - A$ we mean the event formed from $I$ by omitting $A$ and its value. We have thus reduced our problem to finding a good estimate of the latter probability. The earlier procedure—looking for irrelevant attributes—is immediately applicable again. In this way, we can achieve a substantial increase in the quality of the estimate. Note that we do not require that the attributes be considered in any particular order.

# 4    JUSTIFICATION AND EXTENSIONS

The success of our technique depends on whether the vocabulary of attributes used to define events really reflects the way the world works. For example, the attributes $A_1 = day\ of\ the\ week$ and $A_2 = the\ weather$ seem to be fairly independent of each other; there are many contexts where just one of these is relevant. On the other hand, consider $A'_1$, which tells us the day of the week if it is Monday or Wednesday and the weather otherwise, and $A'_2$ which tells us the weather on Monday or Wednesday, and the day otherwise. It is not easy to imagine a context where just one of $\{A'_1, A'_2\}$ is relevant. Both these sets of attributes are equally informative for describing what the world is actually like. Our judgment that $\{A_1, A_2\}$ is better seems to be based on knowledge we have about the causal structure of the world.[7] Our technique is a framework that allows such knowledge to be usefully incorporated into the decision making process. Equally important is that we do not rely on a precise or accurate statement of this knowledge.

Sometimes we have additional knowledge, beyond just a feeling about what a suitable attribute vocabulary should be. A feature of our technique is that it allows easy extensions to cope with many types of extra information. For instance:

- If we are able to provide actual probabilities directly, there is nothing preventing them being used; the estimation procedure can be bypassed when not needed. Similarly, the method can be modified to incorporate statistical data from diverse external sources.

- If we know that some particular attributes can be ignored in certain contexts, then the hypothesis test is redundant and can be omitted. In particular, if some restrictions on the possible relationships between attributes are given to the agent (for example, as a reference hierarchy), this can be used in our process. By avoiding the hypothesis

---

[5]Note that if we did not reuse data, the proof of these theorems would be nearly trivial (because then the hypothesis test would be certain to be independent of both $\hat{p}$ and $\hat{p}_1$). Furthermore, the results would be somewhat tighter; e.g., the composite estimate would be unbiased even for finite sample sizes.

[6]This is not quite correct as stated, because $N$ can grow without bound even as some $N_i$ stays small. The convergence we have in mind here is that, as $N$ tends to infinity, each $N_i$ must be bounded below by $N^c$ for some $c > 0$.

[7]This is reminiscent of the well-known "grue/bleen" paradox ([Goodman, 1955]), which concerns the attribute vocabulary appropriate for inductive inference.

test, we gain computational efficiency and eliminate the chance of error.

- Suppose we know that if some attribute is relevant, then it must affect probabilities in a particular way. As an example, I am not sure whether the probability of traffic congestion on a particular highway depends on which day of the week it is. I know that, if the day of the week is in fact relevant, then this probability is lower on weekends. This knowledge suggests using a different test for independence. We test the hypothesis that the probability is independent of the day, against the alternatives (which have different probabilities for each day, but definitely lower on Saturday and Sunday). We omit details of such a test here. In general, whenever our knowledge can restrict the possible alternatives, a hypothesis test can achieve the same confidence using less data.

- So far, we have regarded the classes considered for pooling as being implicitly defined by the attribute vocabulary. However, our knowledge about the domain may suggest other classes as well. In our earlier example, to estimate the chance of heavy traffic congestion on a Tuesday it might be useful to consider pooling data over just weekdays (Monday to Friday), as well as over the class of all days. It is even possible to use statistical procedures to suggest useful classes, on the basis of previously collected data (but then we must be careful to use a different set of data for the hypothesis test and estimation procedure, because the results would be statistically invalid otherwise).

Finally, we note that the correctness of our technique depends on the stability over time of the underlying probability distribution. If this cannot be assumed, it would be sensible to ignore or discount older observations. There are several standard ways this might be done. Nevertheless, it is clear that robustness against changes in the underlying distribution can only be obtained at the price of slower or less accurate learning.

## 5   COMPARISON TO OTHER WORK

Although many techniques of decision theory have been utilized in artificial intelligence, the decision matrix paradigm of separating the probabilities of states from the utilities has been relatively ignored. Many researchers who adopt the concept of maximizing expected utility (see [Etzioni, 1989b; Horvitz, 1988; Wellman, 1990; Russel and Wefald, 1988]) compute the expected utility for each action directly. A separate computation of utilities has the major advantage of allowing additional information about utilities and probabilities, arising from different sources, to be used.

The description of a state may be detailed enough so that the utility of an action at that state can be computed using knowledge about causality that the agent might have. For example, the agent might know that when the state of the world is such that there is light traffic and the weather is good, then the action of driving ten miles on a freeway must take about ten minutes, because the average velocity in those conditions is 55 miles per hour. Also, by dividing the estimation process into two stages, more historical data will be usable. For example, the agent might conclude that the exact day of the week (say Friday) is relevant to the probability of having heavy traffic, and will therefore use only the historical data about Friday to compute it. But heavy traffic also occurs on other days (although with different probabilities), so that the agent will be able to use all that additional data to compute the expected driving time given heavy traffic. This leads to more accurate estimates.

While some research in AI has adopted this separation of probabilities and utilities (notably [Haddawy and Hanks, 1990; Simon and Kadane, 1975]), the problem of estimating the probabilities in the face of too much initial information has not, to our knowledge, been attacked directly. Some works [Simon and Kadane, 1975] simply assume that the probabilities are known in advance. Others (e.g. [Bundy, 1984; Lee and Mahajan, 1988]) suggest the concept of sampling, but do not discuss which class to sample. Rendell [Rendell, 1983] deals with the concept of sampling on different classes, but in the very limited context of search trees. Etzioni's work [Etzioni, 1989a] on estimating utilities is based on machine learning techniques, which attempt to discover classes over which the utility is homogeneous. Such homogeneity usually arises due to a deterministic relationship between the properties of the class and the utility (such as the driving time given light traffic described above). These techniques do not carry over to estimating probabilities, because the only way to achieve homogeneity in classes of binary values is to have the entire class be all zeros or all ones. I.e., the class will be such that it deterministically forces the truth value of the event. Typically, it is impossible to find an attribute language precise enough to define such classes.

A different approach to finding the right class for estimating probabilities is to treat the problem of inferring the independence structure as a separate task. For example, [Fung and Crawford, 1990] use techniques similar to ours—classical statistics, and in particular, the $\chi^2$ test—in a system which infers qualitative structure from data, modeling this structure as a *probabilistic network*. Once constructed, the network can be used for several purposes, such as estimation. The major drawback of this technique is that a separate data set is required for the construction of the network. It is not clear how this technique can be safely extended to reuse data. Therefore, larger amounts of data will be needed. Our approach is also better in situations where new data is being constantly accumulated, because the new information could cause us to change our decision as to the relevance of certain attributes.

[Fung and Crawford, 1990] also show how to find the smallest possible set of relevant attributes. This procedure is computationally expensive and relies on strong assumptions about the relationships among the attributes. These assumptions also prevent the procedure in [Fung and Crawford, 1990] from being efficiently used in our framework, because each decision situation will need to be investigated separately. This eliminates the computational advantage of computing the entire independence structure simultaneously. Our approach eliminates the attributes one by one, in an arbitrary order. While not guaranteed to find the minimal set of relevant attributes, this technique is much faster and requires no assumptions.

We conclude this section by comparing our methods to the Bayesian approach. It should be noted that, if prior probabilities are available, Bayesian updating (see [Jaynes, 1968]) can, in a sense, replace the $\chi^2$ test described in Section 3. We have chosen not to assume the existence of prior probabilities, and therefore use a technique from classical statistics. A work similar in outlook to ours, which deals with a different problem using Bayesian techniques is Pearl's work about hierarchies of hypotheses [Pearl, 1986].

## 6    CONCLUSION

We have investigated the problem of estimating the conditional probability of a state given some initial information $I$, based on a database of observations. This problem is straightforward when there is plenty of data. However, in many situations, there is little data that exactly matches $I$. Our main result discusses how to utilize the available data in order to decide which attributes in $I$ are irrelevant, and how to use the information about irrelevance to improve the estimate's quality. A feature of our approach is that it uses all the available data for both this decision and for the actual estimation.

We have discussed in detail the assumptions that are required to make our approach sound. For example, the model is simplified by the (commonly made) assumption we call *complete observability*, that is, that all data points contain observations about every attribute. However, we discuss a relaxation of this, *independent observability*, which is far more realistic yet still permits efficient estimation.

The idea behind our approach to estimation is not limited to finding conditional probabilities. For instance, in the context of decision theory which motivates the work in this paper, another important application would be the estimation of utilities; we believe that this is likely to be a straightforward extension of the present work.

The most important factor in the success of our approach will be the quality of the attribute vocabulary that the agent uses to describe the world. Whether made explicit or not, this theme recurs throughout AI. In our approach, this issue is prominent, and we may hope that our more technical results and discussion will serve to shed some light on this fundamental issue.

Finally, we note that one of the advantages of our framework is that it can be extended in several directions, to make use of other knowledge aside from raw observational data. A few suggestions towards this were described in Section 4, but clearly this does not exhaust all the possibilities.

### Acknowledgments

The authors would like to thank Joseph Halpern for comments and discussions relating to this paper.

Some of this work was executed while both authors were employed at IBM Almaden Research Center, 650 Harry Road, San Jose, California 95120-6099. The first author is also supported by an IBM graduate fellowship.